\documentclass[10pt,twocolumn,letterpaper]{article}

\usepackage{cvpr}
\usepackage{times}
\usepackage{epsfig}
\usepackage{graphicx}
\usepackage{amsmath}
\usepackage{amssymb}
\usepackage{pifont}
\usepackage[sort,nocompress]{cite}

\usepackage[utf8,utf8x]{inputenc} 
\usepackage[T1]{fontenc}    
\usepackage{url}            
\usepackage{booktabs}       
\usepackage{amsfonts}       
\usepackage{nicefrac}       
\usepackage{microtype}      
\usepackage{tabu}           
\usepackage{multirow}
\usepackage{subcaption}
\usepackage{bigstrut}
\usepackage{graphicx}
\usepackage{amsmath}
\usepackage{algorithm}
\usepackage{algorithmicx}
\usepackage{algpseudocode}
\usepackage{amsmath,amssymb}
\usepackage[dvipsnames]{xcolor}
\usepackage{subcaption}

\usepackage[dvipsnames]{xcolor}

\newcommand{\cmark}{\ding{51}}%
\newcommand{\xmark}{\ding{55}}%

\usepackage[pagebackref=true,breaklinks=true,letterpaper=true,colorlinks,bookmarks=false]{hyperref}

\cvprfinalcopy 

\def\cvprPaperID{1078} 

\ifcvprfinal\fi
\begin{document}

\title{Global-Local Bidirectional Reasoning for Unsupervised Representation Learning of 3D Point Clouds}

\author{Yongming Rao\textsuperscript{1,2,3},  Jiwen Lu\textsuperscript{1,2,3}\thanks{Corresponding author}
, Jie Zhou\textsuperscript{1,2,3,4}\\
\textsuperscript{1}Department of Automation, Tsinghua University, China\\
\textsuperscript{2}State Key Lab of Intelligent Technologies and Systems, China\\
\textsuperscript{3}Beijing National Research Center for Information Science and Technology, China\\
\textsuperscript{4}Tsinghua Shenzhen International Graduate School, Tsinghua University, China \\
{\tt\small raoyongming95@gmail.com;  \{lujiwen, jzhou\}@tsinghua.edu.cn} \\
}

\maketitle
\thispagestyle{empty}

\begin{abstract}
    Local and global patterns of an object are closely related. Although each part of an object is incomplete, the underlying attributes about the object are shared among all parts, which makes reasoning the whole object from a single part possible. We hypothesize that a powerful representation of a 3D object should model the attributes that are shared between parts and the whole object, and distinguishable from other objects. Based on this hypothesis, we propose to learn point cloud representation by bidirectional reasoning between the local structures at different abstraction hierarchies and the global shape without human supervision. 
    Experimental results on various benchmark datasets demonstrate the unsupervisedly learned representation is even better than supervised representation in discriminative power, generalization ability, and robustness. We show that unsupervisedly trained point cloud models can outperform their supervised counterparts on downstream classification tasks. Most notably,  by simply increasing the channel width of an SSG PointNet++\footnote{Single-Scale Grouping PoinetNet++~\cite{qi2017pointnet++}.}, our unsupervised model surpasses the state-of-the-art supervised methods on both synthetic and real-world 3D object classification datasets. We expect our observations to offer a new perspective on learning better representation from data structures instead of human annotations for point cloud understanding.\footnote{Code:~\href{https://github.com/raoyongming/PointGLR}{https://github.com/raoyongming/PointGLR}}
\end{abstract}

\section{Introduction}

\begin{figure}
\centering
\includegraphics[width=1\linewidth]{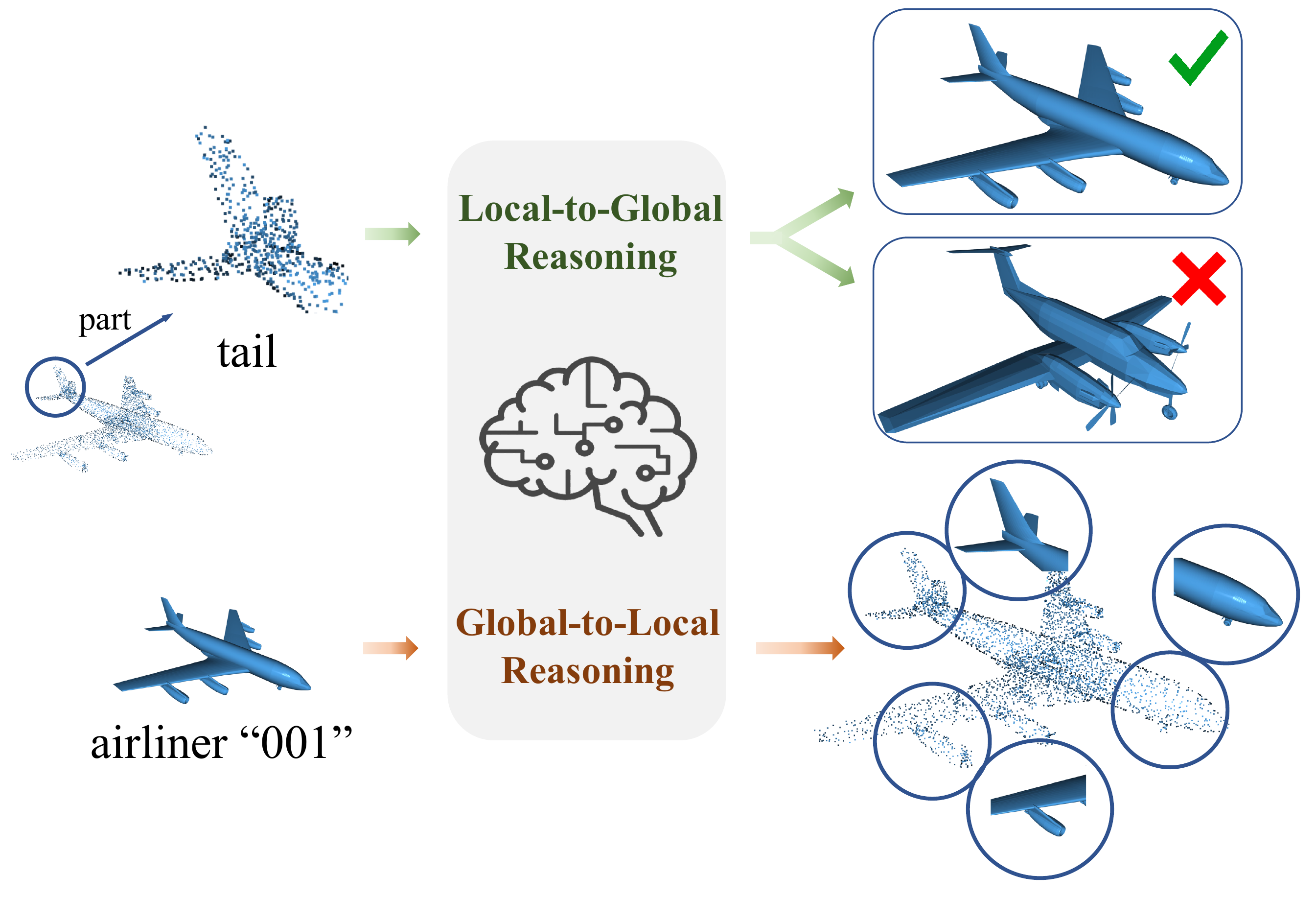}
\vspace{-20pt}
\caption{\small \textbf{Illustration of our main idea.} We propose to learn representation \emph{unsupervisedly} from data structures by training the networks to solve two problems: reasoning the whole object from a single part and reasoning detailed structures from the global representation.}
\label{fig:intro}
\vspace{-15pt}
\end{figure}

Facilitating machines to understand the 3D world is crucial to many important real-world applications, such as autonomous driving, augmented reality and robotics. One core problem on 3D geometric data such as point clouds is learning powerful representations that are discriminative, generic and robust. To tackle this problem, current state-of-the-arts on point cloud analysis~\cite{li2018point, qi2017pointnet++, li2018pointcnn, rscnn-liu2019relation, wu2019pointconv, dgcnn, xu2018spidercnn, atzmon2018point, thomas2019kpconv} are established with the help of extensive human-annotated supervised information. However, manually labeled data require the high cost of human labor and may limit the generalization ability of the learned models. Therefore, unsupervised learning is an attractive direction to obtain generic and robust representations for 3D object understanding.

Learning useful representations from unlabeled data is a fundamental and challenging problem for point cloud analysis. While several efforts have been devoted to learn representation of a point cloud without human supervision~\cite{yang2018foldingnet, LGAN-achlioptas2017learning, deng2018ppf, gadelha2018multiresolution, point-capsule-zhao20193d, valsesia2018learning, L2G-liu2019l2g, han2019multi-angle, li2018point} , these methods are mainly based on self-supervision signals provided by generation or reconstruction tasks, including self-reconstruction~\cite{yang2018foldingnet, LGAN-achlioptas2017learning, deng2018ppf, gadelha2018multiresolution, point-capsule-zhao20193d, valsesia2018learning, li2018point}, local-to-global reconstruction~\cite{L2G-liu2019l2g, han2019multi-angle} and distribution estimation~\cite{LGAN-achlioptas2017learning, li2018point}. These methods have proven to be effective in capturing structural and low-level information of point clouds, but usually fail to learn high-level semantic information from point clouds. Therefore, unsupervised models still perform far behind the state-of-the-art supervised model.  The goal of this work is to explore an unsupervised learning algorithm that can learn both structural information and semantic knowledge to promote the quality of unsupervisedly learned representation.

Different from images where local patches are noisy and usually independent from the whole image (for example, given a patch of a dog, we cannot identify whether this image is about animals or the people nearby), the underlying semantic and structural information is shared in all parts of a 3D object. This distinct property of 3D objects makes reasoning the whole object from a single part possible. Based on this observation, we hypothesize that a powerful representation of a 3D object should model the underlying attributes that are shared between parts and the whole object and distinguishable from other objects. As shown in Figure~\ref{fig:intro}, given a point cloud of a tail of an airplane, a good representation of the tail should reflect the type of the corresponding airplane. Simultaneously, the representation of the whole airplane should contain all the necessary details to infer the local structures of this airplane. 

In this paper, we propose a new scheme for unsupervised point cloud representation learning by bidirectional reasoning between local representations at different abstraction hierarchies in a network and global representation of a 3D object. Our method is simple yet effective, which can be applied to a wide range of deep learning methods for point cloud understanding. While most existing unsupervised learning methods focus on exploiting structure information by learning various autoencoders, our method aims to capture the underlying semantic knowledge shared between local structures and global shape in 3D point clouds. Specifically, the proposed Global-Local Reasoning (GLR) consists of two sub-tasks: 1) local-to-global reasoning: we formulate the problem of capturing shared attributes between local parts and global shape as a self-supervised metric learning problem, where local features are encouraged to be closer to the global feature of the same object than features of other objects, such that the distinct semantic information of each object can be extracted by local representations; 2) global-to-local reasoning: we further use the self-supervised tasks including self-reconstruction and normal estimation to learn global features that contain necessary structural information of 3D objects. 

Our experimental results on several benchmark datasets demonstrate that the unsupervisedly learned point cloud representation is even more discriminative, generalizable and robust than supervised representation in downstream object classification tasks. Our unsupervisedly trained models can consistently outperform their supervised counterparts. With our unsupervised learning method, we show a simple and light-weight SSG PointNet++~\cite{qi2017pointnet++} model can achieve very competitive results with supervised methods  (92.2\% classification accuracy on ModelNet40~\cite{wu2015shapenet}). By simply increasing the channel width, we further obtain 93.0\% and 87.2\% single view accuracy on ModelNet40 and ScanObjectNN~\cite{scanobjectnn-uy2019revisiting} benchmarks respectively, surpassing the state-of-the-art unsupervised and supervised methods, while the supervised version of this model suffers from overfitting.

\section{Related Work}

\begin{figure*}
\centering
\includegraphics[width=1\linewidth]{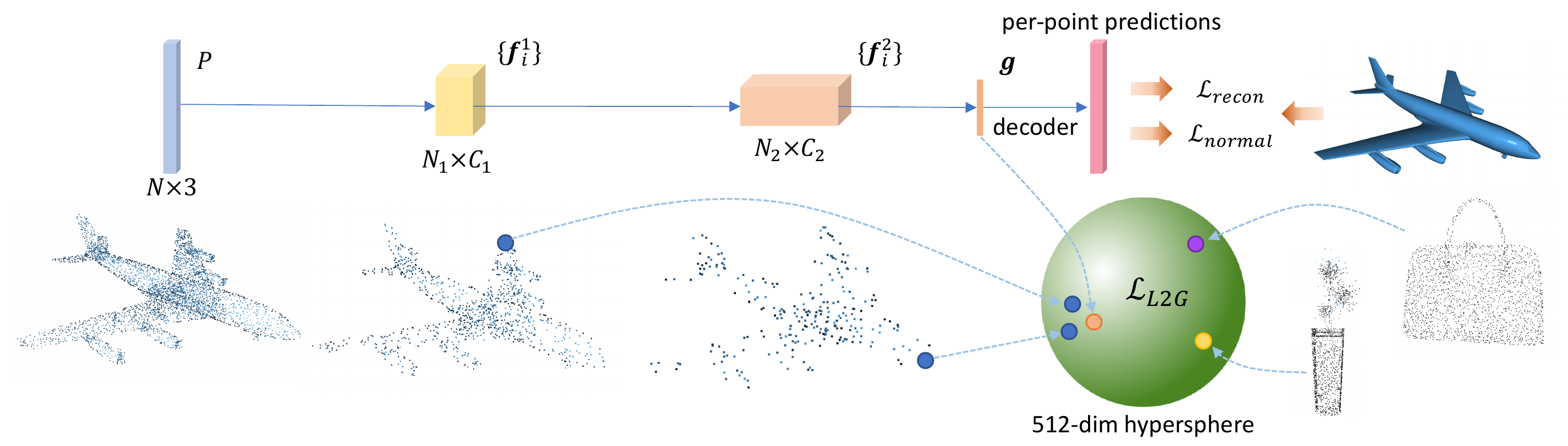}
\vspace{-20pt}
\caption{\textbf{The overall framework of our unsupervised feature learning approach.} The representation is learned by connecting local structures and global shape. We map the local representations at different levels and global representations to shared feature space and use a self-supervised metric learning objective to mine semantic knowledge from data. By further incorporating self-reconstruction and normal estimation tasks, a powerful representation that contains rich semantic and structural information can be learned.}
\label{fig:framework}
\vspace{-10pt}
\end{figure*}

\paragraph{Deep Learning on 3D Point Clouds:} Recent years have witnessed rapid development on 3D point cloud analysis thanks to the deep learning techniques that are designed to consume 3D point clouds directly~\cite{qi2017pointnet, qi2017pointnet++, li2018pointcnn, dgcnn,  rscnn-liu2019relation}. PointNet~\cite{li2018point} pioneers this line of works and designed a deep network that can handle unordered and unstructured 3D points by independently learning on each point and fusing point features with max pooling. Though efficient, PointNet fails to capture local structures, which have proven to be crucial to the success of CNNs. PointNet++~\cite{qi2017pointnet++} is proposed to mitigate this issue by developing a hierarchical grouping architecture to extract local features progressively at different abstraction levels. The subsequent works such as PointCNN~\cite{li2018pointcnn}, PointConv~\cite{wu2019pointconv} and Relation-Shape CNN~\cite{rscnn-liu2019relation} also focus on local structures of point cloud and further improve the quality of captured features. Since only the relation between local and global features is needed, our method is suited for all these PointNet++ variants. While recent works push state-of-the-art of point cloud deep learning by promoting the capacity of networks, this work offers a new route to learn powerful representation in an \emph{unsupervised} fashion, without any human annotations.

\vspace{-10pt}

\paragraph{Unsupervised Representation Learning:} Unsupervised learning has been an important group of methods in computer vision since the earliest day~\cite{fukushima1982neocognitron}, which aims to learn transformations of the data that make the subsequent downstream problem solving easier~\cite{bengio2013representation}. Classical deep methods for unsupervised learning such as autoencoders~\cite{autoencoder-hinton2006reducing}, generative adversarial networks~\cite{goodfellow2014generative} and autoregressive models~\cite{oord2016pixel} learn representation by faithfully reconstructing the input data, which focus on low-level variations in data and is not very useful for downstream tasks like classification. Recent works on self-supervised learning present a powerful family of models that can learn discriminative representations with rich semantic knowledge. This group of methods design various problem generators such that models need to learn useful information from data in order to solve these generated problems~\cite{doersch2015unsupervised,doersch2017multi,cpc-henaff2019data,tian2019contrastive,amdim-bachman2019learning}. In this work, we also follow this line and propose to learn point cloud representation by solving the global-local bidirectional reasoning problem.

There are several prior attempts on learning representation of a point cloud without human supervision~\cite{yang2018foldingnet, LGAN-achlioptas2017learning, deng2018ppf, gadelha2018multiresolution, point-capsule-zhao20193d, valsesia2018learning, L2G-liu2019l2g, han2019multi-angle, li2018point}. These methods discover useful information in the 3D point cloud by performing data reconstruction, which has proven to be effective in learning structural information. However, because of lacking effective semantic supervision, previous methods limit the networks' ability in downstream tasks. Our method resolves this issue by incorporating semantic supervision with structural supervision. With the exploration of high-level semantic knowledge, our method is able to learn discriminative representation like supervised method while maintaining the robustness and generalization of unsupervised representation.

\section{Approach}

The core of 3D point cloud understanding is to learning discriminative, generic and robust representations that can capture the underlying shape. To achieve this goal in an unsupervised manner, we propose to point cloud representation by solving a bidirectional reasoning problem between the local structures and the global shape. The overall framework of our method is presented in Figure~\ref{fig:framework}.

\subsection{Hierarchical Point Cloud Feature Learning}

We begin by reviewing the hierarchical point cloud feature learning framework firstly proposed in PointNet++~\cite{qi2017pointnet++}, on which our method is built. 

Consider a set of 3D points $P \subset \mathbb{R}^3$ with $N$ elements, in which each point $p_i$ is represented by a 3D coordinate. To learn features based on these 3D coordinates, PointNet~\cite{qi2017pointnet} proposes to use a symmetric function $f$ that is invariant to point permutations to transfer point set into feature space:
\begin{equation}
    f(P) = \mathcal{A}(h(p_1), h(p_2), ..., h(p_N)),
\end{equation}
where $h$ is a multi-layer perceptron network that processes each point independently and shares parameters for all points and $\mathcal{A}$ is a symmetric aggregation function like max pooling to summarize features from each point. Since each point is processed independently by $h$, the structural information among points is captured only by the aggregation function $\mathcal{A}$. Therefore, PointNet lacks the ability to capture local context. To address this issue, PointNet++ and its variants~\cite{li2018pointcnn, wu2019pointconv, rscnn-liu2019relation} use a hierarchical structure to learn point cloud feature progressively at different abstraction levels. Specifically, at the $\ell$-th level, point set is abstracted by using iterative furthest point sampling~\cite{qi2017pointnet++} to produce a new set $P^\ell \subset P^{\ell-1}$ with fewer points and we can extract the local geometrical feature $\mathbf{f}_i^{\ell}$ by applying a small PointNet on the local point subset around the centroid for each point $p_i^{\ell} \in P^{\ell}$. The global representation of the point cloud $\mathbf{g}$ is then obtained by applying another small PointNet model on the points and features at the highest abstraction level. 

Almost all previous works~\cite{li2018point, qi2017pointnet++, li2018pointcnn, rscnn-liu2019relation, wu2019pointconv, dgcnn, xu2018spidercnn, atzmon2018point, thomas2019kpconv} on supervised point cloud learning employ an end-to-end training paradigm, where the representation is learned directly from the annotated labels. Although achieved promising performance,  these methods neglect the intrinsic semantic and structure information contained in the point clouds themselves. In this work, we focus on exploring this property of point cloud and provide a very competitive alternative for point cloud representation learning.

To discover the structure and semantic information from data without human annotations, we propose two problems for the networks to solve: \emph{local-to-global reasoning} and \emph{global-to-local reasoning}, which aim to \emph{unsupervisedly} learn semantic and structural knowledge respectively.

\subsection{Local-to-Global Reasoning}

Humans are able to recognize many objects even when only a small part of the object is presented. This fact inspires us to exploit the relation between local parts and global shape as a free and plentiful supervisory signal for training a rich representation for point cloud understanding. Therefore, the goal of local-to-global reasoning is to mine the shared semantic knowledge among different abstraction hierarchies of point clouds. Since global representation usually can better capture the semantic information of 3D objects than local representations, local-to-goal reasoning operates by predicting global representation from local ones. To evaluate the predictions, we formulate the prediction as a self-supervised metric learning problem and use a multi-class N-pair loss~\cite{n-pair-sohn2016improved} to supervise the prediction task. Inspired the idea of instance discrimination~\cite{instance-dis-wu2018unsupervised}, to learn the distinct semantic information for each object, we treat the global representation of the current object as the \emph{positive} sample and use the global representation of other objects as the \emph{negative} samples. In the following, we describe the details of the local-to-global reasoning.

\vspace{-10pt}

\paragraph{Prediction Networks:} Since the local features $\{\mathbf{f}_i^\ell, \forall i,\ell\}$ and global feature $\mathbf{g}$ have different numbers of channels, we cannot directly measure the similarity of them. Thus, we first use prediction networks $\{\phi^\ell, \forall \ell \}$ and $\varphi$ to embed them into a shared feature space, respectively. The prediction networks can be implemented as multi-layer perceptron (MLP) networks and the prediction networks are shared at each abstraction level. 

\vspace{-10pt}

\paragraph{Self-Supervised Metric Learning:} A straightforward method to optimize the predictions is to minimize the absolute overall differences between $\phi^\ell(\mathbf{f}_i^\ell)$ and $\varphi(\mathbf{g})$, i.e. minimize $\sum_{i, \ell} ||\phi^\ell(\mathbf{f}_i^\ell) - \varphi(\mathbf{g})||$. However, this objective may lead to degenerate representations that map all inputs to a constant value. Therefore, we choose to supervise the \emph{relative} quality of the predictions with an unsupervised metric learning task. Specifically, for each embedded local representation $\mathbf{f}_i^\ell$, we enforce its embedding to be closer to the embedded global representation of the same object than any other object. The local-to-global reasoning objective can be written as:
\begin{equation}
    \mathcal{L}_{\text{G2L}}^{i,\ell}  = \log (1 + \sum_{\mathbf{g}_k \neq \mathbf{g}} \exp(s\phi^\ell(\mathbf{f}_i^\ell)^\top \varphi(\mathbf{g}_k) - s\phi^\ell(\mathbf{f}_i^\ell)^\top \varphi(\mathbf{g}))
\end{equation}
and
\begin{equation}
\begin{split}
    \mathcal{L}_{\text{G2L}} & = \frac{1}{M} \sum_{i,\ell}  \mathcal{L}_{\text{G2L}}^{i,\ell} \\
    & = - \frac{1}{M} \sum_{i,\ell} \log \frac{\exp(s\phi^\ell(\mathbf{f}_i^\ell)^\top \varphi(\mathbf{g}))}{\sum_k \exp(s\phi^\ell(\mathbf{f}_i^\ell)^\top \varphi(\mathbf{g}_k))},
\end{split}
\end{equation}
where $\{\mathbf{g}_k, k=1,2,...,m\}$ are the global representations of different point sets in the mini-batch with batch size $m$ and $M$ is the number of local features. Inspired by the studies on metric learning for face recognition~\cite{liu2017sphereface, wang2018cosface, deng2019arcface} that perform metric learning on features on a hypersphere, we normalize the outputs of prediction networks before computing similarities and use a constant value $s=64$~\cite{deng2019arcface} to re-scale the features. Empirically, our experiments show that forcing features to be distributed on a hypersphere with a radius of $s$ will significantly stabilize the training process and improve the discriminative ability of the learned features. 

\vspace{-10pt}

\paragraph{Discussions:} The proposed local-to-global reasoning is connected to mutual information maximization  methods~\cite{cpc-henaff2019data, tian2019contrastive, DeepInfomax-hjelm2018learning, amdim-bachman2019learning} for unsupervised image representation learning. The multi-class N-pair loss can be viewed as a variant of InfoNCE~\cite{cpc1-oord2018representation}. Therefore, minimizing the $\mathcal{L}_{\text{G2L}}$ maximizes the lower bound of the mutual information between local representations and global representation. From this perspective, our method captures the underlying semantic knowledge of a 3D object by maximizing the mutual information of features at different hierarchies. Unlike previous works that performs adversarial learning between the mutual information estimator and the feature encoder~\cite{DeepInfomax-hjelm2018learning} or  maximizes the mutual information of seen patches and unseen patches~\cite{cpc-henaff2019data}, different views of images~\cite{amdim-bachman2019learning} or different modalities of images~\cite{tian2019contrastive} , our work explores the distinct property of point clouds by connecting local and global structures of a 3D object. Furthermore, our local-to-global loss offers a metric learning view of InfoNCE, which is different from previous works that are based on Noise-Contrastive Estimation~\cite{nce-mnih2013learning}. Benefiting our modifications inspired by metric learning and face recognition methods, we observe that our loss is more effective and stable than previous methods on point cloud understanding tasks in our experiments. 

\subsection{Global-to-Local Reasoning}

Since discovering knowledge that is helpful for downstream tasks from unlabeled data is usually quite intractable, local-to-global reasoning may not necessarily lead to useful representations. This fact is also pointed out by studies on mutual information maximization methods~\cite{tian2019contrastive, mi-review}, where evidence shows that larger mutual information may not guarantee a better performance for downstream tasks~\cite{mi-review}. Intuitively, since the local-to-global reasoning only supervises the local representation to be close to the global one, the quality of global representation is critical. This is, if the global representation is well initiated, decent supervision to local representation will be offered, thus creating a \emph{virtuous circle} for the learning of local and global features. On the contrary, the learning process may obtain unpredictable results for the bad initial state of global representation. To avoid this issue, we propose an auxiliary global-to-local reasoning task to supervise the networks for learning useful representation corporately. Specifically, we employ two low-level generation tasks, including self-reconstruction and normal estimation as two self-supervision signals, such that global representation needs to capture the basic structural information of point clouds.

\vspace{-10pt}

\paragraph{Self-Reconstruction:} Self-reconstruction, or point autoencoding, is a widely used technique for unsupervised point cloud representation learning~\cite{yang2018foldingnet, LGAN-achlioptas2017learning, deng2018ppf, gadelha2018multiresolution, point-capsule-zhao20193d, valsesia2018learning, li2018point}. To perform self-reconstruction, we adopt the folding-based~\cite{yang2018foldingnet} decoder $D$ to deform the canonical 2D grid onto 3D coordinates of a point cloud conditioned on the global representation $\mathbf{g}$. The reconstruction error is defined as Chamfer Distance~\cite{chamfer-fan2017point}:
\begin{equation}
     \mathcal{L}_{\text{recon}} = \sum_{p \in P} \min_{x \in D(\mathbf{g})} || x - p ||_2 + \sum_{x \in D(\mathbf{g})}  \min_{p \in P} || x - p ||_2.
\end{equation}

\vspace{-20pt}

\paragraph{Normal Estimation:} Normal estimation is a more challenging task that requires a higher level understanding of the underlying surface information of a 3D shape. Different from previous works~\cite{rscnn-liu2019relation} that pursue the estimation precision, we use this task as a supervisory signal to improve global representation. Thus, we simply concatenate the 3D coordinates with the global representation and employ a shared light-weight MLP $\sigma$ to produce the estimated normals. The cosine loss is used to measure the estimation error:
\begin{equation}
     \mathcal{L}_{\text{normal}} = 1 - \frac{1}{N} \sum_i \cos(\sigma([p_i, \varphi(\mathbf{g})]), p_i^\text{normal}).
\end{equation}

\vspace{-10pt}
Combining the local-to-global reasoning and the global-to-local reasoning, we arrive at the   global-local bidirectional reasoning objective:
\begin{equation}
\mathcal{L}_{\text{GLR}} = \mathcal{L}_{\text{L2G}} + \mathcal{L}_{\text{G2L}} = \mathcal{L}_{\text{L2G}} + \mathcal{L}_{\text{recon}} + \mathcal{L}_{\text{normal}}.
\end{equation}

\subsection{Point Cloud Analysis with GLR}

\paragraph{Unsupervised Learning with GLR:} Point cloud representation can be \emph{unsupervisedly} learned by enforcing networks to solve the proposed global-local reasoning (GLR) problems, where the representation can be used in various downstream point cloud analysis applications like object classification. The quality of unsupervisedly learned representation is usually evaluated by linear separability of  classification task, where a supervised linear SVM~\cite{svm-cortes1995support} model or single-layer neural network is trained on unsupervised representations to measure the test accuracy. For PointNet++~\cite{qi2017pointnet++} model and its variants, we use the aggregated representation for classification task, which is obtained by summarizing embedded global and local representations:
\begin{equation}
\label{eq:agg}
\mathbf{f} = [\mathcal{A}(\{\phi^1(\mathbf{f}_i^1)\}), ..., \mathcal{A}(\{\phi^L(\mathbf{f}_i^L\})), \varphi(\mathbf{g})],
\end{equation}
where we use a max pooling operation $\mathcal{A}$ to aggregate local features of each abstraction level from 1 to $L$ and concatenate these features with the global feature. 

\vspace{-10pt}

\paragraph{Hybrid Learning with GLR:} Since supervisedly learned global representation can be viewed as a good initialization for the proposed GLR framework, our method is also compatible with supervised learning methods, where GLR serves as an auxiliary loss to further improve the robustness of representations. 

\vspace{-10pt}

\paragraph{Implementation:} All of our models is trained on a single GTX 1080ti GPU with deep learning library Pytorch~\cite{pytorch}. To show our method can be used for various point cloud networks, we consider two baseline models:  PointNet++~\cite{qi2017pointnet++} and Relation-Shape CNN (RSCNN)~\cite{rscnn-liu2019relation}. Note that for both baseline models, we use the Single-Scale Grouping (SSG)~\cite{qi2017pointnet++} as the point grouping module, which is more than 3$\times$ smaller than Multi-Scale Grouping (MSG)~\cite{qi2017pointnet++} module used in original PointNet++ model. Besides, we divide the MLP used in each set abstraction layer into two fully connected layers and use them before and after aggregation operation, respectively. Our experiments show this modification can reduce computation and improve performance while keeping the number of parameters unchanged. For unsupervised learning setting, we train a linear SVM~\cite{svm-cortes1995support}  on unsupervised representations of the training data and report the classification accuracy on the test set.  For supervised learning and hybrid learning settings, we use the aforementioned aggregated representation for fair comparison and employ a two-layer classifier where dropout technique~\cite{srivastava2014dropout} with a ratio of 50\% is used for each layer. 
Our models are trained using Adam~\cite{adam-kingma2014adam} optimizer with a base learning rate of 0.001, and we decay the learning rate by 0.7 every 20 epochs. The models are trained for 200 epochs, where the momentum for Batch Normalization~\cite{bn-ioffe2015batch} layers starts with 0.9 and decays with a rate of 0.5 every 20 epochs, following the practice of~\cite{qi2017pointnet++, rscnn-liu2019relation}. Detailed model configurations can be found in Supplementary Material.

\section{Experiments}
We extensively evaluate our method on several widely used point cloud classification benchmark datasets including ModelNet10/40~\cite{wu2015shapenet}, ScanObjectNN~\cite{scanobjectnn-uy2019revisiting} and ScanNet~\cite{dai2017scannet}. We start by evaluating our method on the discriminative power, generalization ability and robustness across datasets and comparing with the state-of-the-art unsupervised and supervised methods. We then provide detailed experiments to analyze our method on model design and complexity. Finally, we visualize the learned representations to have an intuitive understanding of our method. The following describes the details of the experiments, results and analysis.

\subsection{Unsupervised Point Cloud Recognition}

\paragraph{Setups:} We tested our method on ModelNet40/10~\cite{wu2015shapenet} and ScanObjectNN~\cite{scanobjectnn-uy2019revisiting} benchmarks to compare with the state-of-the-arts. ModelNet40 and ModelNet10 comprise 9832/3991 training objects and 2468/908 test objects in 40 and 10 classes respectively, where the points are sampled from CAD models. ScanObjectNN~\cite{scanobjectnn-uy2019revisiting} is a real-world data, where 2902 3D objects are extracted from scans. To conduct cross dataset evaluation, we used the ``object-only'' split in all our experiments. ScanNet~\cite{dai2017scannet} was also used in our cross data evaluation experiments, where we followed the practice in~\cite{li2018pointcnn} to obtain point cloud from indoor scenes according to the instance segmentation labels. In all our experiments, we sample 1024 points for each point cloud for training and evaluation and all our results are measured using a single view without using the multi-view voting trick to show the neat performance of different models. Surface normal information was used to provide unsupervised signals for our models trained on ModelNet and we did not use it as input. For the models trained on ScanObjectNN and ScanNet, we only used the self-reconstruction loss for global-to-local reasoning. 

\begin{figure}
\centering
\includegraphics[width=0.9\linewidth]{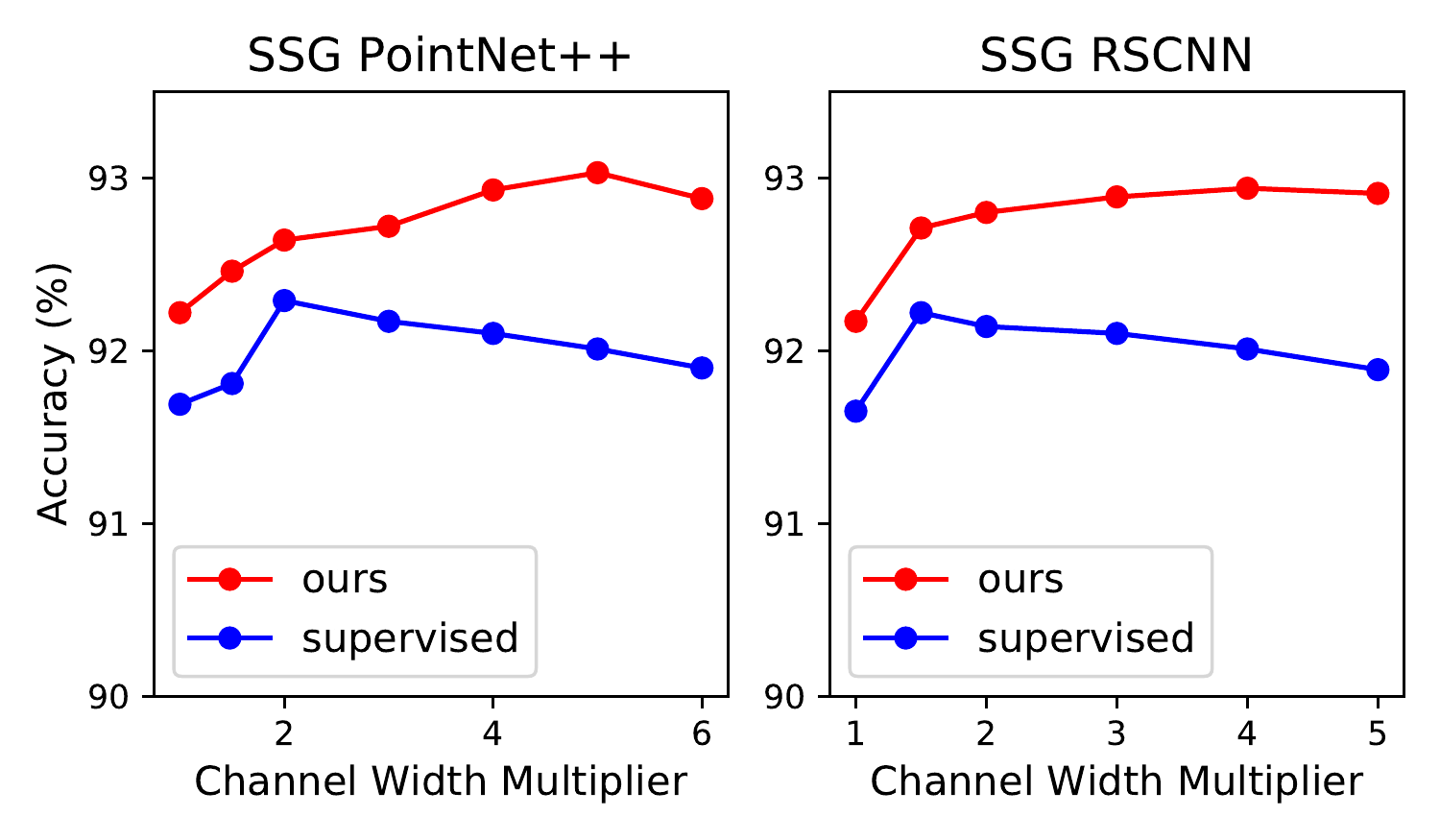}
\vspace{-10pt}
\caption{\small ModelNet40 classification accuracy (\%) of our unsupervised models and their supervised counterparts. }
\label{fig:exp1}
\end{figure}

\begin{table}\small
\centering
\caption{\small Classification accuracy (\%) of three different training strategies on ModelNet40.  }
\label{tab:hybrid}
\vspace{-12pt}
\begin{tabu}to0.48\textwidth{l*{3}{X[c]}}\\
\toprule
  Backbone & \small{Unsupervised} & \small{Supervised} & \small{Hybrid} \\
\midrule
PN++ (Small) & 92.22 & 91.69 & \textbf{92.42}  \\
PN++ (Large) & \textbf{93.02} & 92.01 & 92.76  \\
RSCNN (Small) &  92.17 & 91.65 & \textbf{92.26}  \\
RSCNN (Large) &  \textbf{92.94} & 92.14 &  92.78 \\
\bottomrule
\end{tabu}
\vspace{-5pt}
\end{table}

\vspace{-10pt}

\paragraph{Comparisons with the supervised counterparts:} We first compared our method with the supervised baselines as presented in Figure~\ref{fig:exp1}, where we report the classification accuracy on ModelNet40 using the basic models (1$\times$) and wider models (1.5 to 6$\times$ channel width). Note that we used the same network architecture and training settings for our models and their counterparts for a fair comparison. Clearly, our unsupervised models with different channel widths consistently outperform the supervised counterparts. As increasing the model capacity, our models can achieve better performance and reach the highest accuracy using 5$\times$ PointNet++ and 4$\times$ RSCNN backbones. In the following experiments, we denote the basic 1$\times$ models and the best models as ``Small'' and ``Large'' models respectively.  Besides, we further compared three different training strategies: unsupervised learning, supervised learning and hybrid learning, which are presented in Table~\ref{tab:hybrid}. We see hybrid learning can outperform both supervised and unsupervised models when the networks are small, but the unsupervised method achieves the best performance when large networks are used. We conjecture that the supervised models are prone to overfitting more severely to the training set. All these results reveal that our unsupervised representation is more discriminative and generalizable than its supervised counterpart.

\vspace{-10pt}

\paragraph{Comparisons with the unsupervised state-of-the-arts:} To show the effectiveness of the proposed global-local reasoning method, we compared several variants of our models with the state-of-the-art unsupervised representation learning methods in Table~\ref{tab:unsupervised}. Except for point-based methods, we also compare
with some advanced voxel and view based methods. Note that we only use ModelNet40 as the training data, while some methods are trained on larger ShapeNet~\cite{wu2015shapenet} dataset. Nevertheless, our models outperform all other methods by a large margin. As can be observed, our small PointNet++ model surpasses state-of-the-art methods and our large model significantly advances the best point cloud model (MAP-VAE) by 2.87\% on ModelNet40.

\begin{table}\small \centering
\caption{\small Comparisons of the classification accuracy (\%) of our method against the state-of-the-art \textbf{unsupervised} 3D representation learning methods on ModelNet40 and ModelNet10. $\dagger$ indicates that the model is trained on ShapeNet.}
\label{tab:unsupervised}
\vspace{-12pt}
\begin{tabu}to0.48\textwidth{l*{3}{X[c]}}\\
\toprule
  \multirow{2}{*}{Method} & \multirow{2}{*}{Input} & \multicolumn{2}{c}{Accuracy} \\
  \cmidrule(lr){3-4}& & MN40 & MN10 \\
\midrule
TL Network~\cite{TL-girdhar2016learning} & voxel & 74.40 & -\\
VConv-DAE~\cite{Vconv-dae-sharma2016vconv} & voxel & 75.50 & 80.50 \\
3DGAN~\cite{3dgan-wu2016learning}& voxel & 83.30 & 91.00 \\
VSL~\cite{liu2018learning} & voxel & 84.50 & 91.00 \\
\midrule
VIPGAN~\cite{vipgan-han2019view} & views & 91.98 & 94.05 \\
\midrule
LGAN$^\dagger$~\cite{LGAN-achlioptas2017learning} & points & 85.70 & 95.30 \\
LGAN~\cite{LGAN-achlioptas2017learning} & points & 87.27 & 92.18 \\
FoldingNet$^\dagger$~\cite{yang2018foldingnet} & points & 88.40 & 94.40 \\
FoldingNet~\cite{yang2018foldingnet} & points & 84.36 & 91.85 \\
MRTNet$^\dagger$~\cite{gadelha2018multiresolution} & points & 86.40 & - \\
3D-PointCapsNet~\cite{point-capsule-zhao20193d} & points & 88.90 & - \\
MAP-VAE~\cite{han2019multi-angle} & points & 90.15 & 94.82 \\
\midrule
Ours w/ PN++ (Small) & points & 92.22 & 94.82 \\
Ours w/ PN++ (Large) & points & \textbf{93.02} & \textbf{95.53}\\
Ours w/ RSCNN (Small) & points & 92.17 & 94.60 \\
Ours w/ RSCNN (Large) & points & 92.94 & 95.50 \\
\bottomrule
\end{tabu}
\vspace{-5pt}
\end{table}

\begin{table}\small
\centering
\caption{\small Comparisons of the \emph{single-view} classification accuracy (\%) of our method aganist the state-of-the-art \textbf{supervised} point cloud models on \textbf{ModelNet40}.  We also list results that use more points, normal information (``nor'') or/and multi-view voting trick (``vote'') in {\color{gray} gray} as references. Besides, we show the supervised baselines of our models.  }
\label{tab:supervised-modelnet}
\vspace{-12pt}
\begin{tabu}to0.49\textwidth{l*{3}{X[c]}}\\
\toprule
  Method & \small{\#Points} & \small{Supervised} & \small{Acc.} \\
\midrule
{PointNet~\cite{qi2017pointnet}}  & {1k} & {\cmark} & {89.2} \\
{PointNet++~\cite{qi2017pointnet++}}  & {1k} & {\cmark} & {90.5} \\
\textcolor{gray}{PointNet++~\cite{qi2017pointnet++} (vote)}  & \textcolor{gray}{1k} & \textcolor{gray}{\cmark} & \textcolor{gray}{90.7} \\
SO-Net~\cite{so-net-li2018so} & 1k &  \cmark & 92.5 \\
PointCNN~\cite{li2018pointcnn}  & 1k & \cmark & 92.5 \\
DGCNN~\cite{dgcnn}  & 1k & \cmark & 92.9 \\
DensePoint~\cite{liu2019densepoint} & 1k & \cmark & 92.8 \\
\textcolor{gray}{DensePoint~\cite{liu2019densepoint} (vote)} & \textcolor{gray}{1k} & \textcolor{gray}{\cmark } & \textcolor{gray}{93.2} \\
RSCNN~\cite{rscnn-liu2019relation} & 1k & \cmark & 92.9 \\
\textcolor{gray}{RSCNN~\cite{rscnn-liu2019relation} (vote)} & \textcolor{gray}{1k} & \textcolor{gray}{\cmark } & \textcolor{gray}{93.6} \\
\midrule
\textcolor{gray}{DGCNN~\cite{dgcnn}} & \textcolor{gray}{2k} & \textcolor{gray}{\cmark } & \textcolor{gray}{93.5} \\
\textcolor{gray}{PointNet++~\cite{qi2017pointnet++} (vote, nor)}  & \textcolor{gray}{5k} & \textcolor{gray}{\cmark} & \textcolor{gray}{91.9} \\
\textcolor{gray}{SO-Net~\cite{so-net-li2018so} (nor)} & \textcolor{gray}{5k} & \textcolor{gray}{\cmark } & \textcolor{gray}{93.4} \\
\textcolor{gray}{KPConv~\cite{thomas2019kpconv}} & \textcolor{gray}{$\sim$ 6.8k} & \textcolor{gray}{\cmark } & \textcolor{gray}{92.9} \\
\midrule
PN++ (Large) & 1k & \cmark & 92.1 \\
Ours w/ PN++ (Large) & 1k & \xmark & \textbf{93.0} \\
RSCNN (Large) & 1k & \cmark & 92.0 \\
Ours w/ RSCNN (Large) & 1k & \xmark & 92.9  \\
\bottomrule
\end{tabu}
\end{table}

\begin{table}\small
\centering
\caption{\small Comparisons of the \emph{single-view} classification accuracy (\%) of our method aganist the state-of-the-art \textbf{supervised} point cloud models on \textbf{ScanObjectNN}.  }
\label{tab:supervised-ScanObjectNN}
\vspace{-12pt}
\begin{tabu}to0.48\textwidth{l*{2}{X[c]}}\\
\toprule
  Method & Supervised & Accuracy \\
\midrule
3DmFV~\cite{ben20183dmfv}  & \cmark & 73.8 \\
PointNet~\cite{qi2017pointnet}  & \cmark & 79.2 \\
SpiderCNN~\cite{xu2018spidercnn}  & \cmark & 79.5  \\
PointNet++~\cite{qi2017pointnet++}  & \cmark & 84.3 \\
DGCNN~\cite{dgcnn}  & \cmark & 86.2  \\
PointCNN~\cite{li2018pointcnn}  & \cmark & 85.5  \\
\midrule
Ours w/ PN++ (Large) & \xmark & \textbf{87.2}  \\
Ours w/ RSCNN (Large) & \xmark & 86.9   \\
\bottomrule
\end{tabu}
\vspace{-7pt}
\end{table}

\begin{table}
\small
\centering
\caption{ \small \textbf{Cross dataset evaluation.} We evaluate generalization ability of unsupervised and supervised representations to unseen datasets. We report the classification accuracy (\%) measured using a linear SVM trained on the target dataset. (Sup.: supervised)}
\label{tab:transfer}
\vspace{-12pt}
\begin{tabu}to0.48\textwidth{l*{3}{X[c]}}\\
\toprule
  Task & Sup. & Ours & {$\mathit{\Delta}$} \\
\midrule
ModelNet10 $\to$ ModelNet30 & 85.45 & 92.34 & +6.89 \\
ModelNet30 $\to$ ModelNet10 & 91.32 & 95.47 & +4.15 \\
\midrule
ModelNet40 $\to$ ScanObjectNN & 65.92 & 87.22 & +21.30 \\
ScanObjectNN $\to$ ModelNet40 & 78.76 & 90.80 & +12.04\\
\midrule
ModelNet40 $\to$ ScanNet & 77.31 & 89.23 & +11.92\\
ScanNet $\to$ ModelNet40 & 80.38 & 91.32 & +10.94 \\
\midrule
ScanObjectNN $\to$ ScanNet & 84.31 & 87.96 & +3.63\\
ScanNet $\to$ ScanObjectNN & 82.44 & 85.43 & +2.99\\
\bottomrule
\end{tabu}
\vspace{-5pt}
\end{table}

\vspace{-10pt}

\paragraph{Comparisons with the supervised state-of-the-arts:} More notably, our method can even achieve very competitive results compared to state-of-the-art supervised methods \emph{in an unsupervised manner}. We compared our method with the supervised methods on both the widely used synthetic dataset ModelNet and the recently proposed real-world dataset ScanObjectNN. Our unsupervised representation was trained on ModelNet40 and a linear SVM is then trained on the target dataset to produce predictions. The results are summarized in Table~\ref{tab:supervised-modelnet} and Table~\ref{tab:supervised-ScanObjectNN}. Surprisingly, our unsupervised learned representation can outperform all other state-of-the-arts methods in the single-view setting\footnote{Here we borrow the concept of ``view'' from image recognition literatures, where the number of views represents the number of augmented inputs (\eg rotated or scaled point clouds) used during testing.} on both datasets.  Since only a linear classifier is applied, these results demonstrate that our representation is much more discriminative than the supervised representation on the test set. Moreover, we observe that our representation can achieve very strong results on ScanObjectNN without finetuning. As the categories in ModelNet and ScanObjectNN are different, this evidence indicates that our method can discover semantic knowledge shared in different kinds of objects.

\vspace{-10pt}

\paragraph{Cross Dataset Evaluation:} To further explore the generalization ability of the learned representation, we conducted extensive cross data evaluation experiments on ModelNet, ScanObjectNN and ScanNet, which are varying in categories and sources. Our experiments were conducted based on the unsupervised representations of the PointNet++ large model and we compared the results with the supervised version of this model. Specifically, we trained the features using supervised or unsupervised learning methods on the source dataset and used a linear SVM trained on the target dataset to perform classification. The results are presented in Table~\ref{tab:transfer}, where we used the rest 30 categories in ModelNet40 apart from 10 categories in ModelNet10 to form the ModelNet30 dataset. We see the unsupervisedly learned representation has much stronger transferability than the supervised counterparts and our models generalize well to various unseen data since we learn from \emph{data structures instead of labels}. Our method can maintain strong performance even in cross data evaluation, which reflects the unsupervised representation can be a \emph{generic} representation of 3D objects cross datasets.

\begin{figure}
\centering
\includegraphics[width=0.9\linewidth]{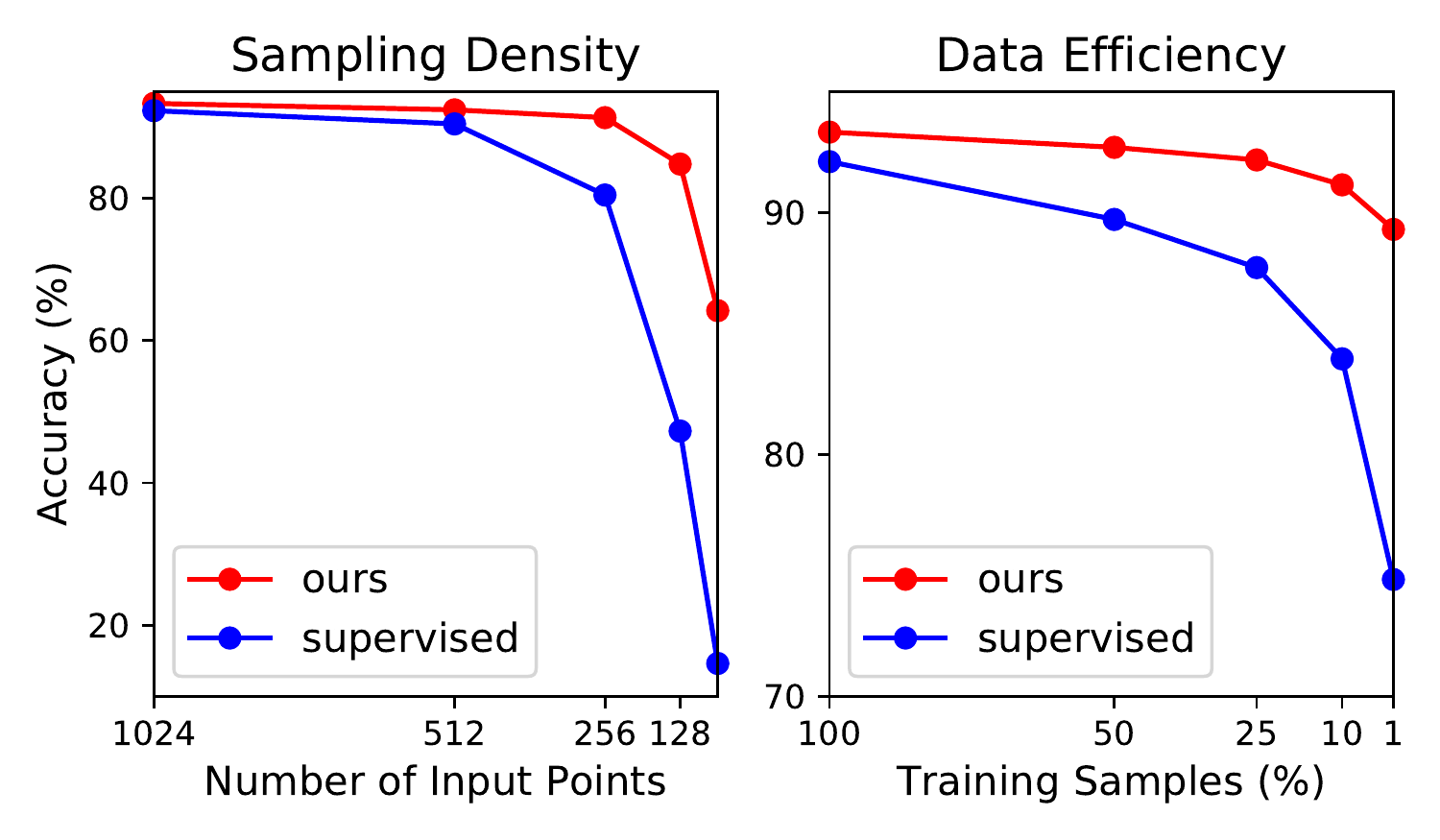}
\vspace{-10pt}
\caption{\small The \textbf{robustness} of our method on sampling density and the number of training samples compared to the supervised baseline.}
\label{fig:robustness}
\vspace{-12pt}
\end{figure}

\vspace{-10pt}

\paragraph{Robustness Analysis:} The robustness of our method on sampling density and the number of training samples is shown in Figure~\ref{fig:robustness}. For the former, we tested the model trained with 1024 points with sparser points of 1024, 512, 256, 128 and 64. Note that different from previous works~\cite{qi2017pointnet++, rscnn-liu2019relation}, we did not perform random input dropout during training. For the latter, we trained the representation with randomly sampled 100\%, 50\%, 25\%, 10\% and 1\% ModelNet40 training set and trained the linear classifier on the whole set. We used the PointNet++ large model in this experiment. Generally, we see our models are much more robust than their supervised versions. Notably, our method can maintain decent performance even when using only 10\% (983 samples) and 1\% (98 samples) training samples and achieve 91.4\% and 89.3\% accuracy on ModelNet40 respectively.

\subsection{Method Design Analysis}

\begin{table}
\small
\centering
\caption{ \small \textbf{Ablation study} of our method. We report the classification accuracy (\%) on ModelNet40. ($\mathcal{L}_{\text{L2G}}$: local-to-global reasoning, $\mathcal{L}_{\text{recon}}$: self-reconstruction, Agg: multi-level feature aggregation in Eq.~(\ref{eq:agg}), $\mathcal{L}_{\text{normal}}$: normal estimation, SN: training on ShapeNet.)}
\label{tab:ablation}
\vspace{-12pt}
\begin{tabu}to0.48\textwidth{l*{6}{X[c]}}\\
\toprule
  Model & $\mathcal{L}_{\text{L2G}}$ & $\mathcal{L}_{\text{recon}}$  & Agg. &  $\mathcal{L}_{\text{normal}}$ & SN & Acc. \\
\midrule
A & & \cmark & & & & 86.77\\
\midrule
B & \cmark & & & &  & 90.02 \\
C & \cmark &  \cmark & & & & 90.96 \\
D & \cmark &  \cmark & \cmark & &  & 91.69 \\
E &  \cmark & \cmark &  \cmark &  \cmark &  & 92.22\\
F &  \cmark & \cmark &  \cmark &  \cmark & \cmark &\textbf{92.30}\\
\bottomrule
\end{tabu}
\end{table}

\begin{table}
\small
\centering
\caption{ \small \textbf{Complexity analysis.} We report the FLOPs and GPU inference throughput with batch size 16. Measured on NVIDIA GTX 1080Ti GPU. (pc/s: point cloud(s) per second) }
\label{tab:complexity}
\vspace{-12pt}
\begin{tabu}to0.48\textwidth{l*{3}{X[c]}}\\
\toprule
  Model & FLOPs & Throughput & Acc. \\
\midrule
MSG PN++~\cite{qi2017pointnet++} & 1.68G & 113pc/s & 90.5 \\
SSG RSCNN \cite{rscnn-liu2019relation} & \textbf{0.30G} & 634pc/s & \textbf{92.2} \\
Our PN++ (Small)  & 0.31G & \textbf{731pc/s} & \textbf{92.2} \\
\midrule
MSG PN++~\cite{qi2017pointnet++} (12 votes) & 14.15G & 9pc/s & 90.7 \\
SSG RSCNN \cite{rscnn-liu2019relation} (10 votes) & \textbf{2.95G} & 63pc/s & 92.7 \\
Our PN++ (Large) & 5.65G & \textbf{194pc/s} & \textbf{93.0} \\

\bottomrule
\end{tabu}
\vspace{-10pt}
\end{table}

\paragraph{Ablation Study:} To examine the effectiveness of our designs, we conducted a detailed ablation study based on the small PointNet++ network. The results are summarized in Table~\ref{tab:ablation}. The baseline model A can be viewed as a variant of FoldingNet~\cite{yang2018foldingnet}, which was trained by self-reconstruction loss only and gets a low classification accuracy of 86.77\%.  We see the model trained by the proposed local-to-global reasoning task (model B) can significantly improve the baseline model by 3.25\%. This convincingly verifies its effectiveness. Then, when incorporating these two losses, the accuracy can be further improved to 90.96\%. We also observe a 0.73\% improvement by aggregating local and global representations (model D). Our full model can be obtained by adding normal estimation supervision (model E), which achieves a notable 92.22\% accuracy on ModelNet40 with a very light-weight network. In addition, we also investigated the training set size by adding more training data (model F) from ShapeNet~\cite{wu2015shapenet}, but obtaining a slight improvement on accuracy (0.08\%). We conjecture that ModelNet is large enough for learning a good representation. Thus we conducted most of the experiments on ModelNet.

\vspace{-10pt}
\paragraph{Complexity Analysis:} Table~\ref{tab:complexity} shows the model complexity in theoretical computation cost (in FLOPs) and actual inference throughput on GPU of our models and several state-of-the-art methods. We see our large model requires considerable computation cost but maintains an acceptable actual cost on GPU due to the simplicity of the SSG model. These results reveal that increasing channel width can achieve a better trade-off on speed and accuracy compared to voting. For computational cost-sensitive applications, we think our learned model can provide strong supervision to train lighter models for real-time applications by model distillation~\cite{hinton2015distilling} or generating pseudo labels~\cite{lee2013pseudo}, which is an interesting direction for future research.

\subsection{Visualization}
\begin{figure}
\centering
\includegraphics[width=0.93\linewidth]{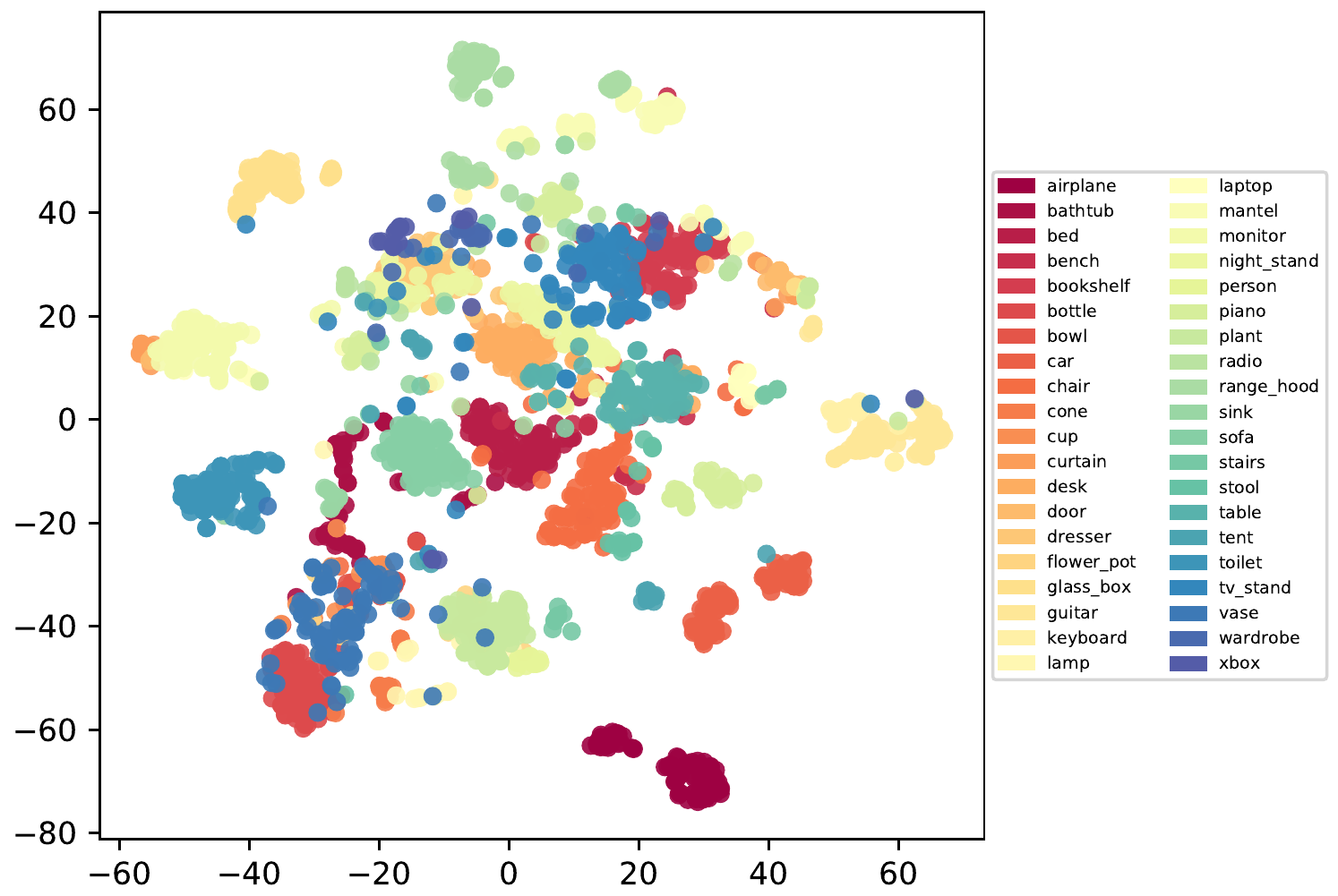}
\vspace{-5pt}
\caption{\small Visualization of unsupervisedly learned representations on the test set of ModelNet40 using t-SNE. Best viewed in color.}
\label{fig:tsne}
\vspace{-10pt}
\end{figure}

\paragraph{Feature Distribution:} To have an intuitive understanding of our models, we visualized the unsupervised learn features on the test set of ModelNet40 in Figure~\ref{fig:tsne}. The features are mapped to 2D space by applying t-SNE~\cite{tsne-maaten2008visualizing}. We see features from different categories are naturally separated without explicit supervision, which reflects the strong discriminative power of our representation.

\begin{figure}
\centering
\includegraphics[width=0.93\linewidth]{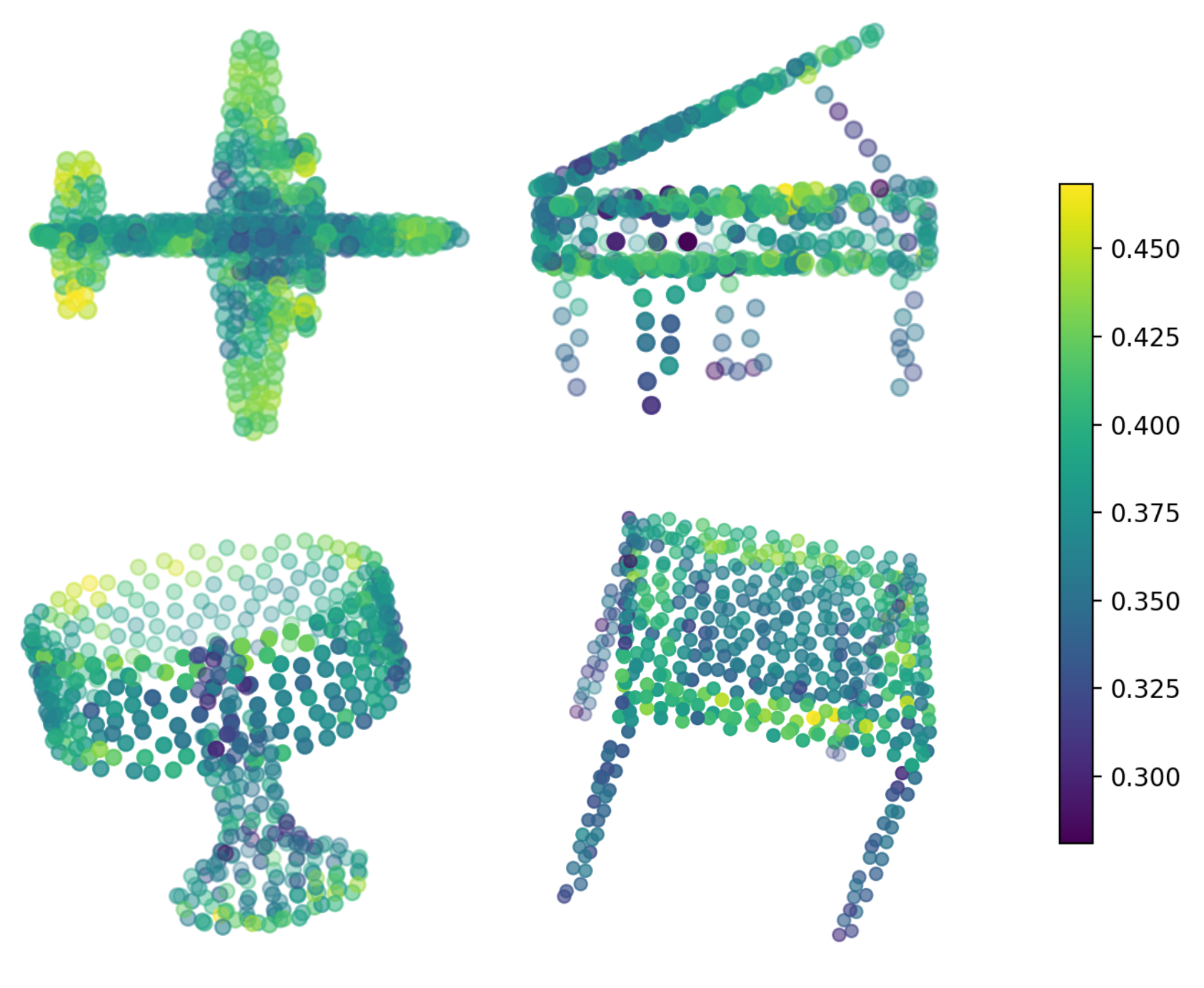}
\vspace{-5pt}
\caption{\small Visualization of the similarity scores between the local features from the first abstraction level and the global feature on the test set of ModelNet40. Best viewed in color.}
\label{fig:glr}
\vspace{-10pt}
\end{figure}

\vspace{-10pt}
\paragraph{Global-Local Relation:} To show the effectiveness of our global-local reasoning method and understand the relation between the global feature and local parts, we visualize the similarity scores between the local features from the first abstraction level and the global feature on the test set of ModelNet40 in Figure~\ref{fig:glr}. We see generally the local features are closed to the global feature (similarity \textgreater 0.25) and the more distinguishable regions usually have higher similarity scores.  

\vspace{-5pt}
\section{Conclusion}
We have proposed a new scheme for unsupervised representation learning of 3D point clouds by bidirectional global-local reasoning. Comprehensive experimental studies have demonstrated our unsupervisedly learned representation can surpass its supervised counterpart and achieve state-of-the-art performance on several widely used benchmarks. We expect our method to open a new door for learning better point cloud representation from data structures instead of human annotation. Transferring the learned knowledge to more efficient models and extending our method to more point cloud analysis scenarios like segmentation and detection are interesting directions in future work. 

\vspace{-5pt}

\section*{Acknowledgement}
This work was supported in part by the National Key Research and Development Program of China under Grant 2017YFA0700802, in part by the National Natural Science Foundation of China under Grant 61822603, Grant U1813218, Grant U1713214, and Grant 61672306, in part by Beijing Academy of Artificial Intelligence (BAAI), in part by a grant from the Institute for Guo Qiang, Tsinghua University, in part by the Shenzhen Fundamental Research Fund (Subject Arrangement) under Grant JCYJ20170412170602564, and in part by Tsinghua University Initiative Scientific Research Program.

\section*{Supplementary Material}
\section*{A. Network Configuration Details}

We first present the details of the single-scale grouping PointNet++~\cite{qi2017pointnet++} used in our experiments. To improve the efficiency of the original PointNet++, we divide
the multi-layer perceptron (MLP) used in each set abstraction (SA) layer of PointNet++ into two fully connected layers and use them before and after the aggregation operation, which can reduce more than 50\% computational cost compared to the original SSG model. The details of the new SSG-SA layer is presented in Table~\ref{tb1}.

\begin{table}[!htbp]\small 
\caption{\small  The detailed architecture of our SSG-SA layer. $C_\text{in}$,  $C_\text{mid}$ and $C_\text{out}$ are the channel widths. $N_\text{in}$ and $N_\text{out}$ are the numbers of input and output points. $K$ is the number of sampled neighboring points.}
\label{tb1}
\begin{tabu}to0.45\textwidth{*{3}{X[c]}}\toprule
input size & layer type &  output size\\
\midrule
($N_\text{in}$, $C_\text{in}$) & Ball Query & ($N_\text{out}$, $C_\text{mid}$, $K$) \\
($N_\text{out}$, $C_\text{mid}$, $K$) & Conv+BN+ReLU & ($N_\text{out}$, $C_\text{mid}$, $K$) \\
($N_\text{out}$, $C_\text{mid}$, $K$) & Max Pooling & ($N_\text{out}$, $C_\text{mid}$) \\
($N_\text{out}$, $C_\text{mid}$) & Conv+BN+ReLU & ($N_\text{out}$, $C_\text{out}$) \\
\bottomrule
\end{tabu}
\end{table}

\begin{table*}[!]
\caption{The detailed architecture of our SSG PointNet++ model and the auxiliary networks with channel width multipler $M$.} 
\label{tb2}
\begin{tabu}to\textwidth{X[c,1.5]X[c,1]X[c,3.5]X[c,1]X[c,1]}\toprule
input & input size  & layer type & output size &  output name\\
\midrule
\multicolumn{5}{l}{\emph{backbone network}} \\
points & (1024, 3) & SSG-SA(512, 48, 0.23, [3, $64M$, $128M$]) & (512, $128M$) & sa1 \\
sa1 & (512, $128M$) & SSG-SA(128, 64, 0.32, [3, $128M$, $512M$]) & (128, $512M$) & sa2 \\
sa2 & (128, $512M$) & MLP([512, 1024]) + Max Pooling & (1, $1024M$) & sa3 \\
\midrule
\multicolumn{5}{l}{\emph{prediction networks}} \\
sa1 & (512, $128M$) & MLP([$128M$, min($128M$, 512), 512]) & (512, 512) & pred1 \\
sa2 & (128, $512M$) & MLP([$512M$, 512, 512]) & (128, 512) & pred2 \\
sa3 & (1, $1024M$) & MLP([$1024M$, 512, 512]) & (1, 512) & pred3 \\
\midrule
\multicolumn{5}{l}{\emph{aggregated representation}} \\
pred1, pred2, pred3 & - & Max Pooling + Concatenation & (1, 1536) & agg  \\
\midrule
\multicolumn{5}{l}{\emph{self-reconstruction networks}} \\
agg & (1, 1536) & MLP([1536 + 2, 512, 256, 3]) & (1024, 3) & recon\_mid \\
agg, recon\_mid & (1024, 1539) & MLP([1536 + 3, 512, 256, 3]) & (1024, 3) & recon \\
\midrule
\multicolumn{5}{l}{\emph{normal estimation networks}} \\
agg, points & (1024, 1539) & MLP([1536 + 3, 512, 256, 3]) & (1024, 3) & normal \\
\bottomrule
\end{tabu}
\end{table*}

For clearness, we use the following notations to describe the layer and corresponding setting format:
\begin{itemize}
    \item SSG-SA($N$, $K$, $r$, [$C_\text{in}$,  $C_\text{mid}$, $C_\text{out}$]) is a single-scale grouping set abstraction layer with $N$ local regions of ball radius $r$ and the number of sampled neighboring points $K$ using channel with configuration [$C_\text{in}$,  $C_\text{mid}$, $C_\text{out}$].
    \item MLP([$C_1$,  ..., $C_d$]) is a $d-1$ layer multi-layer perceptron with channel width $C_1$,  ..., $C_d$.
\end{itemize}

The overall network architecture used in our experiments is summarized in Table~\ref{tb2}, where $M$ is the channel width multiplier. We use the same hyper-parameters of SA layers as~\cite{rscnn-liu2019relation}.

For experiments based on Relation-shape CNN~\cite{rscnn-liu2019relation}, we use the SSG version of this model following the official implementation.

\section*{B. Experiment Details}
\paragraph{ScanNet Experiments:} ScanNet~\cite{dai2017scannet} is a richly annotated dataset of 3D reconstructed meshes of indoor scenes, which contains 1513 scanned and reconstructed scenes. To obtain the 3D object classification dataset from the original ScanNet annotations, we use the instance segmentation labels to extract point clouds of each instances from the complete scenes. We use the ScanNetV2 annotations and splits for training and evaluation, where there are 1201 and 312 scenes for training and 
testing respectively. Following~\cite{li2018pointcnn}, we select objects from 17 categories. 12060 and 3416 objects are extracted as the training and testing sets for object classification task.

\paragraph{Linear SVM:} We use the linear SVM classifier provided by scikit-learn library\footnote{\url{https://scikit-learn.org/stable/modules/generated/sklearn.svm.LinearSVC.html}} to evaluate the unsupervised learning algorithms. We use the default parameters in all our experiments. We only extract one feature for each object to form the training and evaluation data. Data augmentation techniques are not used to train the linear SVM.

{\small
\bibliographystyle{ieee_fullname}
\bibliography{reference}
}

\end{document}